\documentclass[letterpaper, 10 pt, conference]{IEEEtran}

\IEEEoverridecommandlockouts

\usepackage[letterpaper,left=.75in,right=.75in,top=.75in,bottom=.75in]{geometry}

\usepackage{amsfonts}
\usepackage{amsmath}
\usepackage{amssymb}
\usepackage[pdftex]{graphicx}
\usepackage{import}
\usepackage{siunitx}
\usepackage{subcaption}
\usepackage{xcolor}

\usepackage{hyperref}
\hypersetup{
    colorlinks=true,
    linkcolor=black,
    citecolor=black,
    filecolor=black,
    urlcolor=black,
}
\usepackage[T1]{fontenc}
\usepackage[utf8]{inputenc}
\usepackage{csquotes}
\usepackage[english]{babel}
\usepackage[export]{adjustbox}
\usepackage[font=small]{caption}
\usepackage{subcaption}
\usepackage{textcomp}
\usepackage{placeins}

\graphicspath{{images/}}

\newcommand{\norm}[1]{\left\lVert#1\right\rVert}
\newcommand{\abs}[1]{|#1|}
\DeclareMathOperator*{\argmin}{arg\,min}

\title{\vspace{.25in}
Evidential Occupancy Grid Map Augmentation using Deep Learning
\thanks{\textcopyright 2018 IEEE}
}

\author{
\IEEEauthorblockN{Sascha Wirges and Christoph Stiller}
\IEEEauthorblockA{
Mobile Perception Systems Group\\
FZI Research Center for Information Technology\\
Karlsruhe, Germany\\
Email: \{wirges, stiller\}@fzi.de
}
\and
\IEEEauthorblockN{Felix Hartenbach}
\IEEEauthorblockA{
Institute of Measurement and Control\\
Karlsruhe Institute of Technology (KIT)\\
Karlsruhe, Germany\\
Email: uvdsu@student.kit.edu}
}

\begin{document}

\maketitle
\thispagestyle{empty}
\pagestyle{empty}

\begin{abstract}
A detailed environment representation is a crucial component of automated vehicles.
Using single range sensor scans, data is often too sparse and subject to occlusions.
Therefore, we present a method to augment occupancy grid maps from single views to be similar to evidential occupancy maps acquired from different views using Deep Learning.
To accomplish this, we estimate motion between subsequent range sensor measurements and create an evidential 3D voxel map in an extensive post-processing step.
Within this voxel map, we explicitly model uncertainty using evidence theory and create a 2D projection using combination rules.
As input for our neural networks, we use a multi-layer grid map consisting of the three features \textit{detections}, \textit{transmissions} and \textit{intensity}, each for ground and non-ground measurements.
Finally, we perform a quantitative and qualitative evaluation which shows that different network architectures accurately infer evidential measures in real-time.
\end{abstract}
\section{Introduction}
\label{sec:introduction}

For a safe use of mobile robotic systems detailed maps of the environment are required.
However, maps created only from most recent, single measurements are often sparse and subject to occlusions.
In order to gather additional knowledge about the scene further domain knowledge is needed.

There are different ways to accumulate information on the environment.
With odometry estimation one way is to improve object reconstruction by accumulating multiple registered measurements.
However, as the scene is usually dynamic, past sensor information can only be used up to a certain extent.
Another way to augment information is to decompose the environment into independent objects, perform classification and reconstruct these objects using highly accurate ground-truth data (e.g. \cite{Menze2015}).
Although these approaches lead to accurate results for certain object classes, they are computationally expensive, not generic and usually need manually labeled data.

Therefore, our objective in this work is to estimate a generic, precise and augmented environment model using only a single range sensor measurement represented as occupancy grid map.
As there is a large amount of sensor data available, this process should also take advantage of data-driven optimization methods such as Deep Learning.

Our main contribution is to provide a framework for evidential grid map augmentation using Deep Learning.
By estimating motion in an offline process, we are able to create a highly accurate map of the static environment w.r.t. a fixed reference frame.
As the views change due to motion, this map typically becomes denser and contains less occlusions.
We use multiple registered measurements to estimate an evidential 3D occupancy grid map of the environment, perform ground surface segmentation and project the evidential masses onto a planar occupancy grid map using combination rules.
As this approach is computationally expensive it is only performed for generating training data labels.
Using these labels, we then train different neural networks using an occupancy grid map of only one corresponding range sensor measurement as input.
Our method is suitable for learning accurate maps with different neural network architectures.

\begin{figure}[t!]
\centering
\scriptsize
\def\svgwidth{\columnwidth}
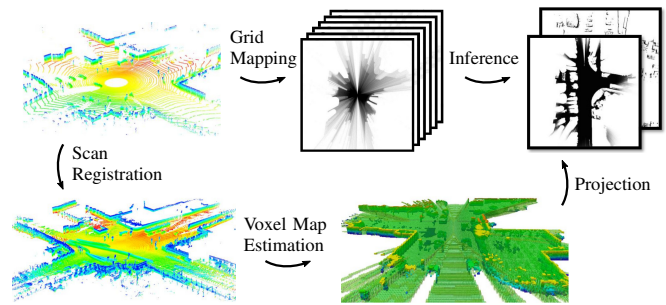
\caption{Occupancy map augmentation processing steps.
We create a multi-layer grid map containing detections, transmissions and intensities from a single range sensor scan which serves as input of the deep inference network.
The inference output is an evidential two-layer occupancy grid map.
The network is trained on projections from an evidential voxel map estimated from registered, subsequent range sensor scans.
}
\label{fig:system_overview}
\end{figure}

At first we review related work on occupancy grid maps and neural network architectures in Section~\ref{sec:related_work}.
We then explain the range sensor data preprocessing to obtain training examples in Section~\ref{sec:preprocessing}.
Then, after recalling the training and validation metrics, we provide information on the training process and network parameters in Section~\ref{sec:training}.
We perform a quantitative and qualitative evaluation of different configurations in Section~\ref{sec:evaluation}.
Finally, we conclude our work and propose future plans for grid map inference in Section~\ref{sec:conclusion}.

\section{Related Work}
\label{sec:related_work}

Occupancy mapping was first developed for planar grids (2D) and later extended to the 3D domain \cite{Elfes1991}.
Today, implementations for both 2D \cite{Fankhauser2016} and 3D \cite{Hornung2013} applications are available open source.
Occupancy grid mapping has applications in collision avoidance \cite{Borenstein1991}, sensor fusion \cite{Thrun1998}, object tracking \cite{Wang2007}, and simultaneous localization and mapping \cite{Hess2016}.
A variety of sensor models suitable for grid maps exist, e.g. correlation-based \cite{Biber2003} or beam-based \cite{Schaefer2017}.

The field of environment augmentation is still ongoing research.
In an offline process Menze et al. fit 3D CAD models to manually labeled vehicles \cite{Menze2015}.
Engelmann et al. perform pose estimation and shape reconstruction with compact shape manifolds on stereo camera images \cite{Engelmann2016}.
Although these approaches yield accurate results, we rate these approaches as not suitable for real-time / online applications as they work only on detected obstacle instances and a computationally expensive optimization is performed.

In the last years, the accuracy of deep convolutional neural networks (CNNs) in image classification, object detection and localization \cite{Russakovsky2015} continuously increased.
Today, CNNs yield the best results in these domains.
In \cite{Ronneberger2015}, Ronneberger et al. generalize the approach of fully convolutional neural networks \cite{Long2015} and outperform previous approaches in cell segmentation tasks.
The network is similar to encoder-decoder models using convolution-pooling layers to increase the receptive field, thus decreasing the spatial resolution.
However, their \textit{Unet} architecture uses skip connections between complementing pooling / unpooling layers to maintain a high spatial resolution.

In \cite{He2015}, the authors show that using a deeper residual network (\textit{Resnet}) the accuracy w.r.t. other architectures can be improved.
The performance of Resnets can then be made more computationally efficient by using 1x1 convolutions \cite{Szegedy2016}.

\section{Preprocessing}
\label{sec:preprocessing}

As depicted in Fig.~\ref{fig:system_overview}, the inference framework consists of several preprocessing steps.
After providing the foundations in sections \ref{subsec:preprocessing_range_sensor_odometry} and \ref{subsec:preprocessing_ground_surface_estimation}, we describe the methods used to create inference input grid maps in Section~\ref{subsec:preprocessing_input_grid_map} and evidential target occupancy grid maps in Section~\ref{subsec:preprocessing_target_occupancy_grid_map}.

\subsection{Range Sensor Scan Registration}
\label{subsec:preprocessing_range_sensor_odometry}

\begin{figure}[h]
\centering
\def\svgwidth{0.95\columnwidth}
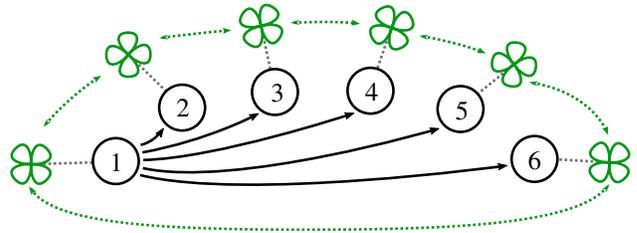
\caption{Per-batch registration scheme.
Each range sensor scan (green) corresponds to a pose relative to the reference pose 1, indicated by black arrows.
We search for closest points between adjacent scans and between the first and the last scan, illustrated by the green arrows.}
\label{fig:registration_scheme}
\end{figure}

In our two-part registration, we represent range sensor scans as point sets.
First, we perform globally consistent generalized Iterative-Closest-Points (GICP) \cite{Segal2009} in batches of six scans in parallel (see Fig.~\ref{fig:registration_scheme}).
Second, we use the resulting pose estimates as observations in a subsequent pose graph optimization, e.g. as described in \cite{Grisetti2010}.
Each points' covariance is estimated based on its 10 nearest neighbors.

\begin{figure}
\centering
\begin{subfigure}[h]{0.49\columnwidth}
\includegraphics[width=\columnwidth]{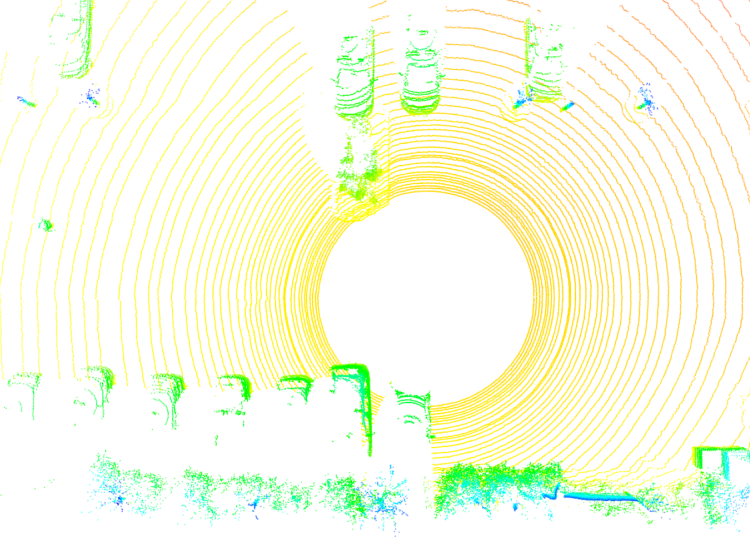}
\caption{Single scan}
\end{subfigure}
\begin{subfigure}[h]{0.49\columnwidth}
\includegraphics[width=\columnwidth]{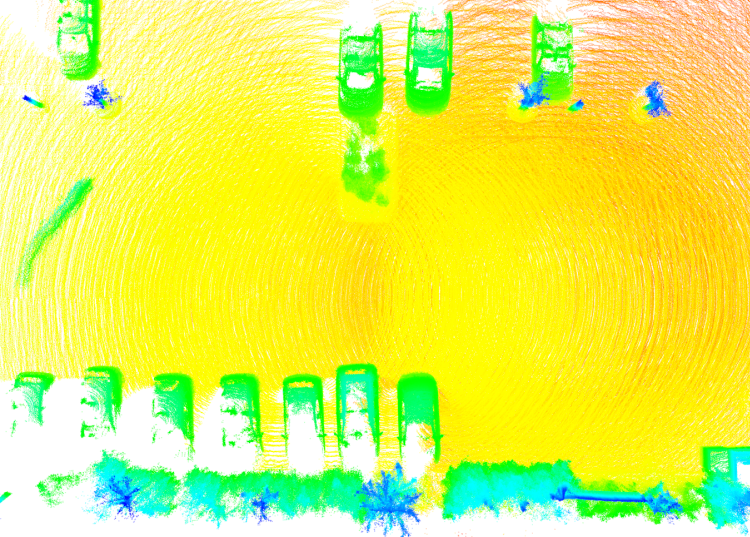}
\caption{41 scans}
\end{subfigure}
\caption{Scene top view (colored by height) measured by one scan and 41 registered scans.
The scene is mainly composed of cars and trees with a pedestrian walking on the left side.
As the pedestrian is moving, they are visible on multiple positions in the registered point set.}
\label{fig:registration}
\end{figure}

Point correspondences between two scans are obtained by nearest neighbor search in the reference frame.
We then keep the reference pose fixed and estimate the remaining poses by minimizing the sum of GICP error functions induced by all point correspondences.
In the pose graph optimization step we insert the pose difference between adjacent scans as observations which results in a multi-edge pose graph to optimize.
Finally, we determine the poses similar to algorithm 2 presented in \cite{Grisetti2010}.
We perform both steps using a nonlinear Least-Squares solver \cite{ceres}.
An exemplary registration result compared to a single range sensor scan is depicted in Fig.~\ref{fig:registration}.

\subsection{Ground Surface Estimation}
\label{subsec:preprocessing_ground_surface_estimation}

In the majority of scenarios we observed the ground surface to be flat.
Therefore, we make a plane assumption in this work using either one scan or an accumulation of multiple registered scans as input.
We perform nonlinear Least-Squares optimization \cite{ceres} to find the optimal plane parameters
\begin{equation}
\textbf{pl}^* = \argmin_{\textbf{pl}} \sum_{\mathbf{p} \in \mathcal{P}} \rho\left( \norm{\mathbf{e} \left( \textbf{pl}, \mathbf{p} \right)}^2 \right)
\end{equation}
which minimize the accumulated point-to-plane error for all points $\mathbf{p}$ of the point set, where $\mathbf{e} \left( \textbf{pl}, \mathbf{p} \right)$ denotes the distance vector between $\mathbf{p}$ and its plane projection point.
The loss function $\rho$ is chosen to be the Cauchy loss with a small scale (5 cm) to strictly robustify against outliers.
In addition, we remove all points far below the estimated ground plane as they are likely a result of multipath propagation.

\subsection{Input Grid Map}
\label{subsec:preprocessing_input_grid_map}

Given a single range sensor scan, we perform ground surface estimation as described in Section~\ref{subsec:preprocessing_ground_surface_estimation}.
We obtain a \textit{ground} and a \textit{non-ground} subset of points which are then used to compute a grid map.
Each grid cell contains the number of \textit{detections}, free-space \textit{transmissions} and the average reflected energy, termed as \textit{intensity}.
We determine these values by casting rays to all detected end points from the sensor origin.
Fig.~\ref{fig:non_ground_intensities}--\ref{fig:ground_transmissions} show the resulting six grid map layers in an exemplary scenario.

\subsection{Target Occupancy Grid Map}
\label{subsec:preprocessing_target_occupancy_grid_map}

\begin{figure}
    \centering
    \begin{subfigure}[h]{0.49\columnwidth}
        \includegraphics[width=\columnwidth]{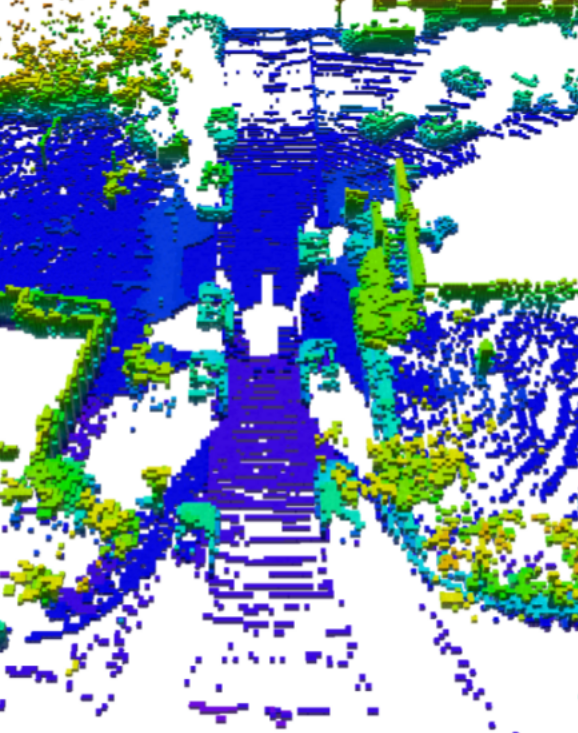}
        \caption{Complete}
    \end{subfigure}
    \begin{subfigure}[h]{0.49\columnwidth}
        \includegraphics[width=\columnwidth]{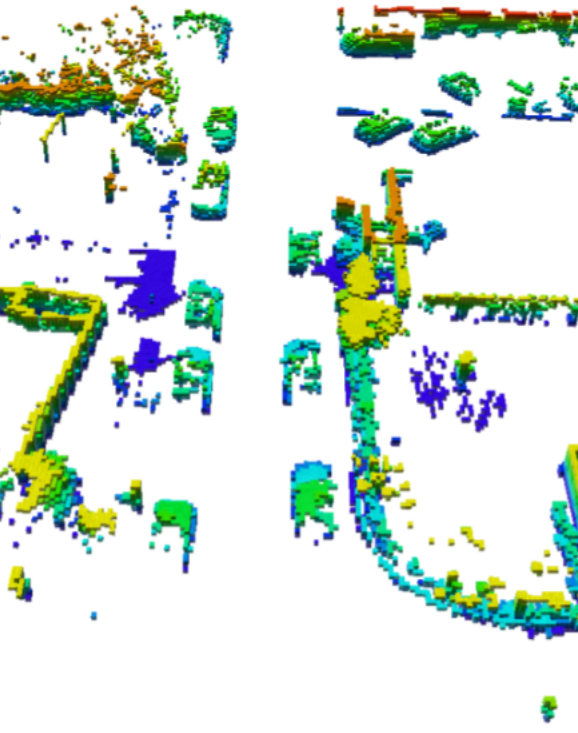}
        \caption{Driving corridor}
    \end{subfigure}    
    \caption{Evidential voxel map before and after vehicle driving corridor segmentation.
    Color indicates the height above ground.}
    \label{fig:voxel_map}
\end{figure}

For each scan, we take all scan data in the time interval up to $\pm 2$s to create an evidential 3D voxel map.
Instead of using data in a time interval one could also use data within a range of poses.
Compared to a single measurement, the scene has an extended field of view and is less subject to occlusions.
Here, we implicitly make the assumption that the world is static which leads to artifacts when obstacles (e.g. cars, pedestrians) are moving.
However, evidential combination of registered measurements can partially mitigate this problem, as moving objects generate higher per-cell uncertainties which is presented in the following.

Given a set of subsequent, registered scans, we estimate the ground surface the same way as for the input.
The scans are inserted into a 3D voxel map similar to \cite{Hornung2013} by raycasting to the scans' endpoints from the sensor origins.
We then reduce the 3D voxel map to the vehicle's driving corridor between 0.2 m and 3 m above ground by performing the ground surface segmentation described in Section~\ref{subsec:preprocessing_ground_surface_estimation}.
Fig.~\ref{fig:voxel_map} depicts the voxel maps before and after driving corridor segmentation for one scenario.

Each voxel includes the number of reflections and transmissions.
For each voxel only two hypotheses, $O$ for a cell being occupied, or $F$ for a cell being free, are possible which results in the frame of discernment $\Theta = \left\lbrace O, F \right\rbrace$.
The beliefs
\begin{align}
\text{bel}(O) &= 1 - \text{pl}(F) = e(\left\lbrace O \right\rbrace) \\
\text{bel}(F) &= 1 - \text{pl}(O) = e(\left\lbrace F \right\rbrace)
\end{align}
of a cell being occupied or free can also be expressed by their plausibility counterparts $\text{pl}(\cdot)$ and depends on the elementary evidences reflections $e_R$ and transmissions $e_T$, respectively.

Based on the recommendation for their inverse sensor model in \cite{Hornung2013}, we choose the elementary evidences as
\begin{align}
e_R(\left\lbrace O \right\rbrace) &= 0.4, &e_T(\left\lbrace O \right\rbrace) &= 0 \\
e_R(\left\lbrace F \right\rbrace) &= 0, &e_T(\left\lbrace F \right\rbrace) &= 0.1, \\
e_R(\left\lbrace \Theta \right\rbrace) &= 0.6, &e_T(\left\lbrace \Theta \right\rbrace) &= 0.9,
\end{align}
so that they lead to similar behavior when combination rules are applied.

These evidences are combined using Yager's Rule of Combination, in the following indexed by the $\cup$ symbol.
This results in the combined evidence of a voxel $V$
\begin{align}
e_{\cup, V}(\left\lbrace O \right\rbrace) 					&= (1 - e_R(\left\lbrace \Theta \right\rbrace)^m) \cdot e_T(\left\lbrace \Theta \right\rbrace)^n \\
e_{\cup, V}(\left\lbrace F \right\rbrace) 					&= (1 - e_T(\left\lbrace \Theta \right\rbrace)^n) \cdot e_R(\left\lbrace \Theta \right\rbrace)^m \\
e_{\cup, V}(\left\lbrace \Theta \right\rbrace) 				&= 1 - e_{\cup, V}(\left\lbrace O \right\rbrace) - e_{\cup, V}(\left\lbrace F \right\rbrace) \\
e_{\cup, V}(\left\lbrace \emptyset \right\rbrace) 			&= 0
\end{align}
with the number of reflections $m$ and the number of transmissions $n$.
As a result of reflections and transmissions within the same voxel, conflicting evidence masses are assigned to the entire frame of discernment.
Thus grid cells temporarily covered by moving objects yield a high uncertainty which mitigates the assumption of a static environment for target data generation.

Finally, the evidential 3D voxel map is projected onto a plane in order to get an evidential 2D occupancy grid map.
In the following, we summarize $k = 1 \ldots K$ voxels $V^{(k)}$ on top of each other to a \textit{pillar} $P$.
However, the voxel evidences cannot be combined as shown previously as they describe different locations.
As for the voxels, a pillar is assigned the two hypotheses \textit{O} and \textit{F}.
Whereas evidence for an occupied voxel is also evidence that the corresponding pillar is occupied, evidence for a voxel being free is less an evidence for the whole pillar being free.
The latter is only the case if all pillar voxels have high evidence for being free.
On the one hand, this yields the belief
\begin{equation}
\text{bel}_{\cup, P}(F) = e_{\cup}(\left\lbrace F \right\rbrace) = \prod_{k=1}^{K} e_{\cup, V}^{(k)}(\left\lbrace F \right\rbrace)
\end{equation}
for a pillar being free, similar to chaining all voxel evidences for being free by logical \textit{and} functions.
On the other hand, the belief
\begin{equation}
\text{bel}_{\cup, P}(O) = e_{\cup}(\left\lbrace O \right\rbrace) = 1 - \prod_{k=1}^{K} 1 - e_{\cup, V}^{(k)}(\left\lbrace O \right\rbrace),
\end{equation}
for a pillar being occupied can be interpreted as chaining all voxel evidences for not being free by logical \textit{or} functions.

As a result, the belief
\begin{equation}
\text{bel}_{\cup, P}(\Theta) = 1 = e_{\cup, P}(\Theta) + e_{\cup, P}(F) + e_{\cup, P}(O)
\end{equation}
shows that the correct hypothesis is in $\Theta$ with evidence
\begin{equation}
e_{\cup, P}(\Theta) = 1 - e_{\cup, P}(F) - e_{\cup, P}(O).
\end{equation}

Finally, we obtain a two-channel grid map used for training our inference networks containing the beliefs $\text{bel}_{\cup, P}(O)$ and $\text{bel}_{\cup, P}(F)$.
Any additional channel would be redundant.

\section{Training}
\label{sec:training}

\subsection{Data Set and Training Strategy}
We created a dataset containing 7707 range sensor scans using a Velodyne HDL64E-S2 lidar.
The dataset contains sequences from different traffic scenarios such as parking lots, highways, cities or rural roads.
After preprocessing the data (see Section~\ref{sec:preprocessing}), we split the dataset into 5995 training and 712 evaluation samples covering different driving environments.
We created grid maps with an initial size of 100\,m\,$\times$\,100\,m and quadratic cells with an edge length of 12.5 cm.
To further increase the number of training examples, we applied random rotation and offset from the sensor origin and cropped areas of 64\,m\,$\times$\,64\,m that were then used as training examples.
Due to our limited computational resources we trained all networks using Minibatch-SGD with four samples per batch, layer normalization \cite{Ba2016} and used the Adam optimizer with a learning rate of $1\cdot10^{-4}$ for Unets and $5\cdot10^{-4}$ for Resnets.

\subsection{Metrics}
\label{subsec:training_metrics}

To train the networks, we use cell-wise metrics.
Given a grid with I cells, we define the loss
\begin{equation}
L = \left(\sum_{i = 1}^{I} w^{(i)}\right)^{-1} \sum_{i = 1}^{I} w^{(i)} l^{(i)}
\label{eq:loss}
\end{equation}
as the average weighted per-cell loss with the belief $\text{bel}(O)$ for a cell being occupied, $\text{bel}(F)$ for the cell being free and the estimates $\text{bel}^\prime(O)$ and $\text{bel}^\prime(F)$, respectively.

Using the above notation, we define the per-cell residuals
\begin{align}
\label{eq:res_o}
\epsilon_O &= \text{bel}(O) - \text{bel}^\prime(O), \\
\label{eq:res_f}
\epsilon_F &= \text{bel}(F) - \text{bel}^\prime(F)
\end{align}
and per-cell $L_1$ and $L_2$ losses
\begin{align}
\label{eq:l_1}
l_1 &= \abs{\epsilon_O} + \abs{\epsilon_F}, \\
\label{eq:l_2}
l_2 &= \epsilon_O^2 + \epsilon_F^2.
\end{align}

As the target data includes inaccuracies due to registration errors, sensor noise or moving obstacles, we want to scale the per-cell loss depending on the target data uncertainty.
The loss definitions (Eq. \ref{eq:l_1}, \ref{eq:l_2}) with $w = 1$ would therefore yield to an approximation of the target data uncertainty.
To make the inference result more independent to this uncertainty, we propose two modifications of the above cost terms.

First, we propose to scale the per-cell loss depending on the weight
\begin{equation}
w_k = 1 + k \cdot (C - 1) \qquad k \in \left[ 0, 1 \right]
\label{eq:W-k}
\end{equation}
which depends on the target data certainty
\begin{equation}
C = \text{bel}(O) + \text{bel}(F).
\end{equation}
Second, we suggest to adapt the cost asymetrically for false free predictions, e.g. in the $L_1$ loss function such that
\begin{equation}
l_{1, k} = \abs{\epsilon_O} + \abs{\epsilon_F} + k \cdot \epsilon_F \qquad k \in \left[ 0, 1 \right]
\label{eq:F-k}
\end{equation}
yields to an underestimation of $\text{bel}(F)$.

\subsection{Networks and Hyperparameters}
Our Unets are a generalized modification of \cite{Ronneberger2015}.
The Resnets used are similar to \cite{He2015} but use dilated convolutions \cite{Yu2015}.
As depicted in Table~\ref{tab:abbreviations}, we explicitly set the network hyperparameters by varying the loss function type, the initial number of filters, the stack size and the network depth.
A stack is defined as consecutive convolutions and ReLUs with identical filter size.
After each stack, pooling is performed (Unet) or the dilation rate gets doubled (Resnet) and for each the filter size gets doubled.
The Network depth describes the number of stacks of the encoder part.
A bottom stack follows as well as a decoder part with the same number of stacks as the encoder.
Thus, a network with Depth D has 2D+1 Stacks.
The maximum number of convolutions CL of a network path is greater or equal (2D+1)S.
The number of filters cannot be altered in residual blocks so extra 1x1 convolutions are added between the stacks.

\begin{table}[t]
\centering
\begin{tabular}{l|l}
\textbf{Net} 			& Architecture and ID (U: UNet, Res: Resnet) \\
\hline
\\
\textbf{HP Explicit} 	& Explicitly set hyper parameters \\
\hline
\textbf{Loss} 			& Loss function definition \\
\textbf{F} 				& Number of filters in the first layer \\
\textbf{S} 				& Stack size (Convolutions + ReLUs) \\
\textbf{D} 				& Network depth \\
\\
\textbf{HP Implicit} 	& Implicitly set hyper parameters \\
\hline
\textbf{CL} 			& Maximum number of convolution layers \\
\textbf{Rec} 			& Receptive field size in multiples of 0.125 m \\
\textbf{Para} 			& Total number of parameters \\
\\
\textbf{Training} 		& Training parameters \\
\hline
\textbf{Ep} 			& Number of epochs \\
\textbf{Step} 			& Number of training steps \\
\\
\textbf{Metrics} 		& Evaluation metrics \\
\hline
\textbf{L1} 			& Standard $L_1$ loss (Eq. \ref{eq:l_1}, $w = 1$) \\
\textbf{L2} 			& Standard $L_2$ loss (Eq. \ref{eq:l_2}, $w = 1$) \\
\textbf{RelUnc} 		& Relative uncertainty (Eq. \ref{eq:rel_unc}) \\
\textbf{False O} 		& False occupied (Eq. \ref{eq:false_o}) \\
\textbf{False F} 		& False-free (Eq. \ref{eq:false_f}) \\
\textbf{Time} 			& Inference time \\
\end{tabular}
\caption{Abbreviations used in training and evaluation.}
\label{tab:abbreviations}
\end{table}

We trained Unets and Resnets with several configurations.
One Unet (U5, \textit{NoSplit}) was trained for a three channel grid map input without splitting the range sensor scans into ground and non-ground points.
Besides the standard loss (Eq.~\ref{eq:loss}) with $w=1$, we also train one Unet (U6, \textit{W0.9}) with the weight definition according to Eq.~\ref{eq:W-k} with $k=0.9$ depending on the target data certainty.
Unet U7 (\textit{F0.8}) was trained with the asymetric loss in Eq.~\ref{eq:F-k} with $k=0.8$.
\section{Evaluation}
\label{sec:evaluation}

\sisetup{table-number-alignment=center, exponent-product=\cdot, output-decimal-marker = {.}}
\begin{table*}[t]
\centering
\begin{tabular}{r|lrrr|rrr|rr|rrrrrr}
 & \multicolumn{4}{c}{\textbf{HP Explicit}} & \multicolumn{3}{c}{\textbf{HP Implicit}} & \multicolumn{2}{c}{\textbf{Training}} & \multicolumn{6}{c}{\textbf{Metrics}}\\
\textbf{Net} & \textbf{Loss} & \textbf{F} & \textbf{S} & \textbf{D} & \textbf{CL} & \textbf{Rec} & \textbf{Para} & \textbf{Ep} & \textbf{Step} & {\textbf{L1}} & {\textbf{L2}} & \textbf{False O} & \textbf{False F} & \textbf{RelUnc} & \textbf{Time}\\
& & & & & & & $10^3$ & & $10^3$ & $10^{-4}$ & $10^{-5}$ & & $10^{-2}$ & $10^{-2}$ & ms \\
\hline
U1 				& L1 			& 8 	& 3 & 3 & 25 & 152 	& 123 	& 40 	& 30 	& 8.52 			& 5.18 			& 5.03			& 44.80 		& 1.02 			& \textbf{38} \\
U2				& L1 			& 16 	& 3 & 3 & 25 & 152 	& 487 	& 40 	& 60 	& 8.14 			& 4.97 			& 5.62			& 36.60 		& 1.03 			& 44 \\
U3 				& L1 			& 8 	& 3 & 4 & 32 & 313 	& 492 	& 40 	& 60 	& 8.44 			& 5.19 			& 5.13			& 49.30 		& 1.01 			& 39 \\
U4 				& L2 			& 8 	& 3 & 3 & 25 & 152 	& 123 	& 40 	& 60 	& 10.40 		& \textbf{4.35} & 68.10			& 59.50 		& 1.00 			& \textbf{38} \\
U5 				& L1 NoSplit 	& 8 	& 3 & 3 & 25 & 152 	& 123 	& 40 	& 60 	& 8.93 			& 5.64 			& 6.53			& 55.30			& 1.03 			& \textbf{38} \\
\textbf{U6} 	& L1 W0.9 		& 8 	& 3 & 3 & 25 & 152 	& 123 	& 40 	& 60 	& 11.30 		& 6.57 			& 42.20			& 83.40 		& \textbf{0.92} & \textbf{38} \\
\textbf{U7} 	& L1 F0.8 		& 8 	& 3 & 3 & 25 & 152 	& 123 	& 40 	& 60 	& 11.20 		& 7.74 			& \textbf{5.26}	& \textbf{3.42} & 1.13 			& \textbf{38} \\
Res1			& L1 			& 8 	& 2 & 3 & 26 & 93 	& 73 	& 120 	& 181 	& 8.18 			& 4.97 			& 4.21 			& 7.11 			& 1.02 			& 72 \\
Res2			& L1 			& 8 	& 2 & 4 & 36 & 193 	& 295 	& 87 	& 130 	& 8.13 			& 4.89 			& 6.22 			& 10.10 		& 1.01 			& 119 \\
Res3			& L1 			& 8 	& 1 & 4 & 22 & 105 	& 172 	& 120 	& 180 	& 8.16 			& 5.03 			& 3.23 			& 6.20 			& 1.02 			& 82 \\
\textbf{Res4}	& L1 			& 16 	& 2 & 3 & 26 & 93 	& 288 	& 103 	& 150 	& \textbf{8.08} & 4.93 			& 3.47 			& 7.21 			& 1.02 			& 107
\end{tabular}\\
\caption{Selection of network parameters and evaluation results.
We evaluated Unets and dilated Resnets with different hyperparameters.
Networks for qualitative analysis and best performance results highlighted.
}
\label{tab:quantitative_results}
\end{table*}

Table~\ref{tab:quantitative_results} depicts the best network configurations we found in our evaluation.
For the abbreviations used, please refer to Table~\ref{tab:abbreviations}.

\subsection{Quality Metrics}
\label{subsec:quality_metrics}

As we infer scalar beliefs of cells being free or occupied, common metrics used in the evaluation of binary classifiers, e.g. precision or recall \cite{Powers2011} cannot be applied.
Therefore, we define further metrics to evaluate the inference quality in addition to the metrics presented in Section~\ref{subsec:training_metrics} on which the networks are trained on.

We define the relative uncertainty
\begin{equation}
m_{\text{RelUnc}} = \frac{\hat{U}}{U} = \frac{1 - \text{bel}^\prime(F) - \text{bel}^\prime(O)}{1 - \text{bel}(F) - \text{bel}(O)}
\label{eq:rel_unc}
\end{equation}
as the ratio of the predicted uncertainty $\hat{U}$ and the target uncertainty $U$.
In the default case, $m_{\text{RelUnc}} \approx 1$.
However, $m_{\text{RelUnc}}$ might vary a lot for some configurations and helps us describe their influence.

Whereas $L_1$ and $L_2$ norm are assembled by per-channel distances, e.g. $\text{bel}^\prime(F)$ and $\text{bel}(F)$, we aim to penalize contradictory inference w.r.t. the target data.
Therefore, we define the false occupied~/~free metrics
\begin{align}
\label{eq:false_o}
m_{\text{FalseO}} &= \text{max}(0, \text{bel}^\prime(O) + \text{bel}(F) - 1), \\
\label{eq:false_f}
m_{\text{FalseF}} &= \text{max}(0, \text{bel}(O) + \text{bel}^\prime(F) - 1)
\end{align}
that penalize areas of inferred high belief in contradiction to the target data.
Compared to $m_{\text{FalseO}}$, $m_{\text{FalseF}}$ is usually more critical because an acute collision might be possible if obstacles are detected too small or not detected at all.
For both of the metrics above, areas of high uncertainty, either labeled or predicted, will induce only little error.

\subsection{Processing Time}
We evaluated the processing times on a 2.5 GHz six core Intel Xeon E5-2640 CPU with 15 MB cache and an NVIDIA GeForce GTX 1080 Ti GPU with 11 GB graphics memory.
The input grid map estimation including ground plane segmentation takes 18 ms on average.
Given this input, we evaluated the inference times for different networks and hyperparameter configurations.
The results are depicted in Table~\ref{tab:quantitative_results}.
We observe the Unets to be faster than Resnets as they might take advantage of higher parallelization.
However, even for Resnets we achieve real-time performance (processing time less than 100 ms) for one configuration.

\subsection{Discussion}

\begin{figure*}[t]
\centering
\begin{subfigure}[h]{0.16\linewidth}
\includegraphics[width=\linewidth]{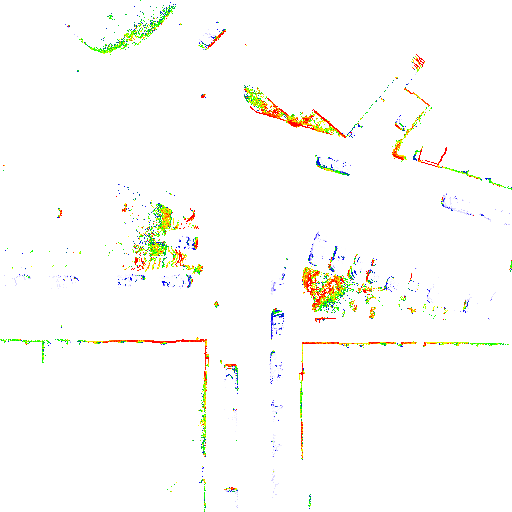}
\caption{Non-gr. intens.}
\label{fig:non_ground_intensities}
\end{subfigure}
\begin{subfigure}[h]{0.16\linewidth}
\includegraphics[width=\linewidth]{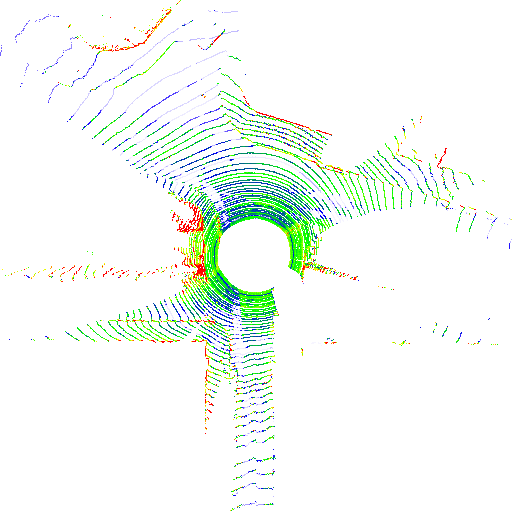}
\caption{Ground intensities}
\label{fig:ground_intensities}
\end{subfigure}
\begin{subfigure}[h]{0.16\linewidth}
\includegraphics[width=\linewidth]{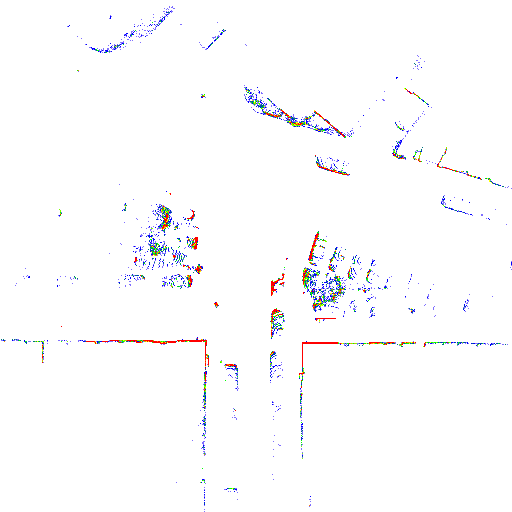}
\caption{Non-gr. detections}
\label{fig:non_ground_detections}
\end{subfigure}    
\begin{subfigure}[h]{0.16\linewidth}
\includegraphics[width=\linewidth]{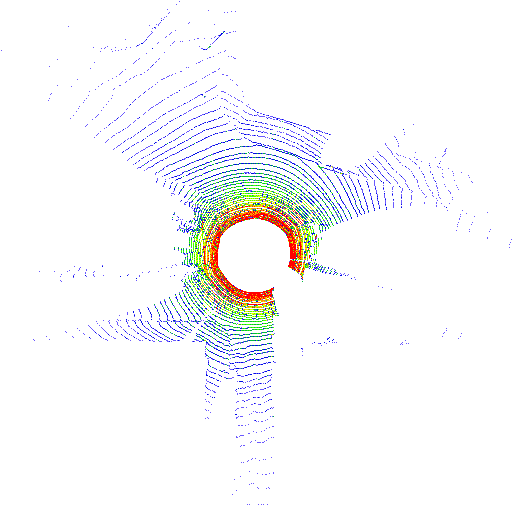}
\caption{Ground detections}
\label{fig:ground_detections}
\end{subfigure}
\begin{subfigure}[h]{0.16\linewidth}
\includegraphics[width=\linewidth]{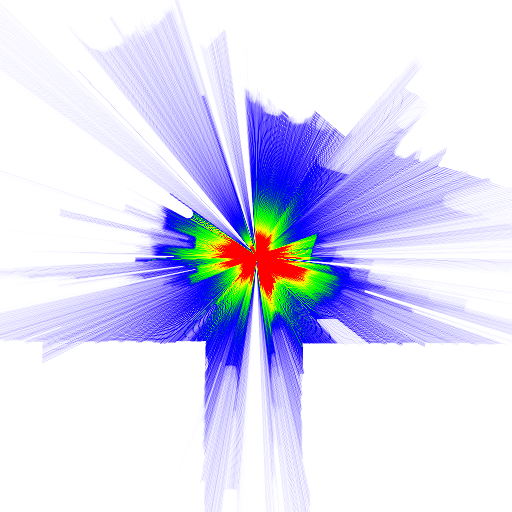}
\caption{Non-gr. transmiss.}
\label{fig:non_ground_transmissions}
\end{subfigure}
\begin{subfigure}[h]{0.16\linewidth}
\includegraphics[width=\linewidth]{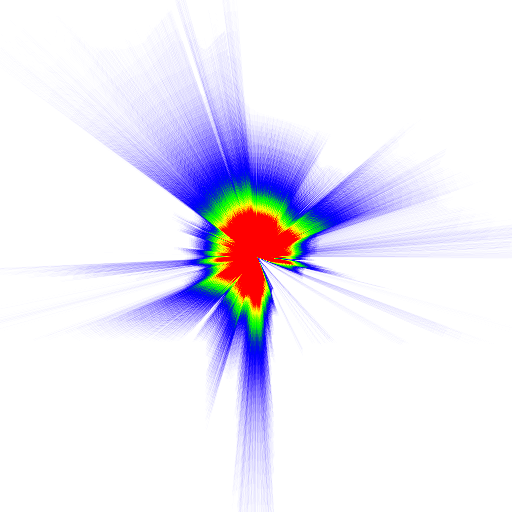}
\caption{Ground transmiss.}
\label{fig:ground_transmissions}
\end{subfigure}

\begin{subfigure}[h]{0.24\linewidth}
\includegraphics[width=\linewidth]{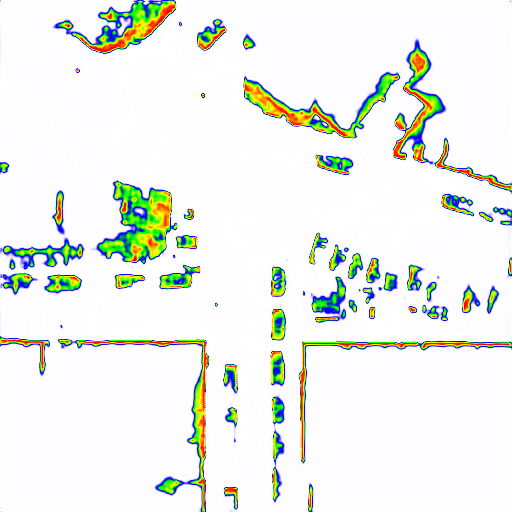}
\end{subfigure}
\begin{subfigure}[h]{0.24\linewidth}
\includegraphics[width=\linewidth]{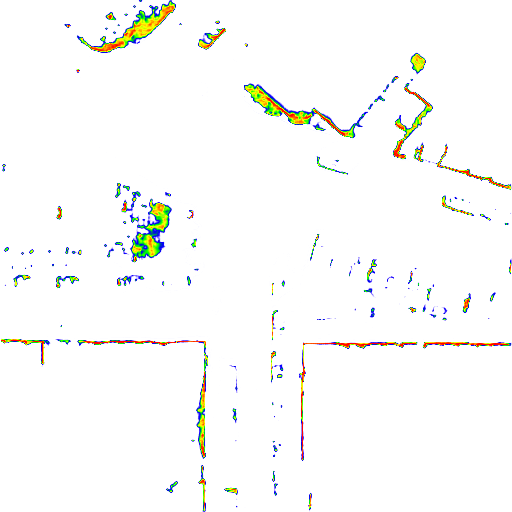}
\end{subfigure}
\begin{subfigure}[h]{0.24\linewidth}
\includegraphics[width=\linewidth]{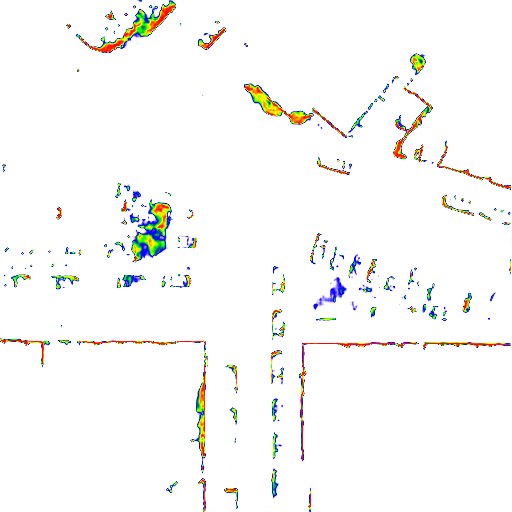}
\end{subfigure}
\begin{subfigure}[h]{0.24\linewidth}
\includegraphics[width=\linewidth]{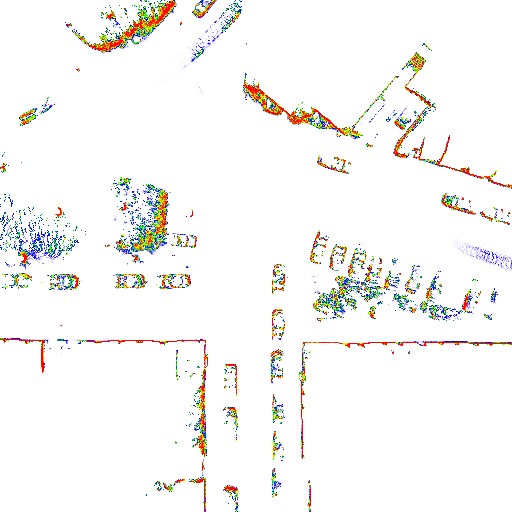}
\end{subfigure}

\begin{subfigure}[h]{0.24\linewidth}
\includegraphics[width=\linewidth]{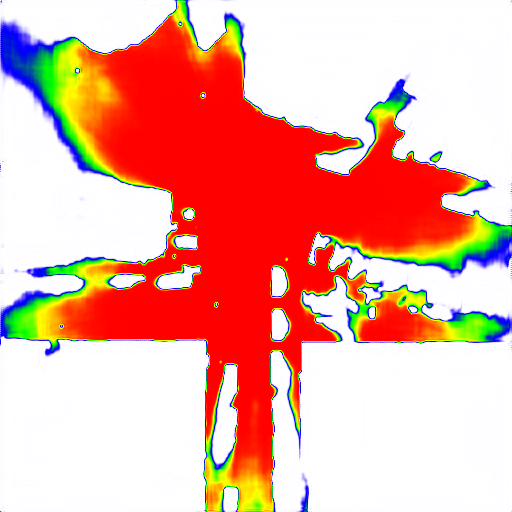}
\caption{U6}
\label{fig:results_u6}
\end{subfigure}    
\begin{subfigure}[h]{0.24\linewidth}
\includegraphics[width=\linewidth]{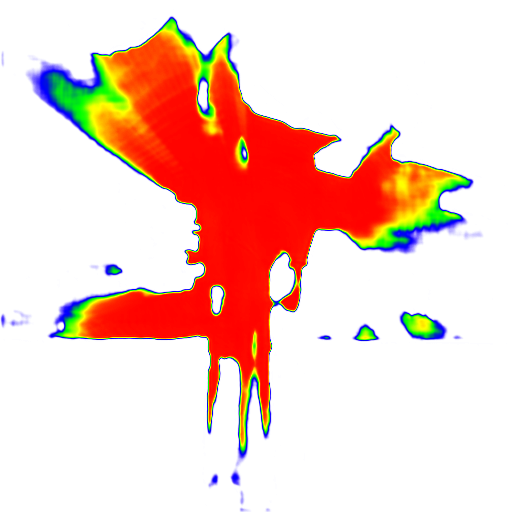}
\caption{U7}
\label{fig:results_u7}
\end{subfigure}
\begin{subfigure}[h]{0.24\linewidth}
\includegraphics[width=\linewidth]{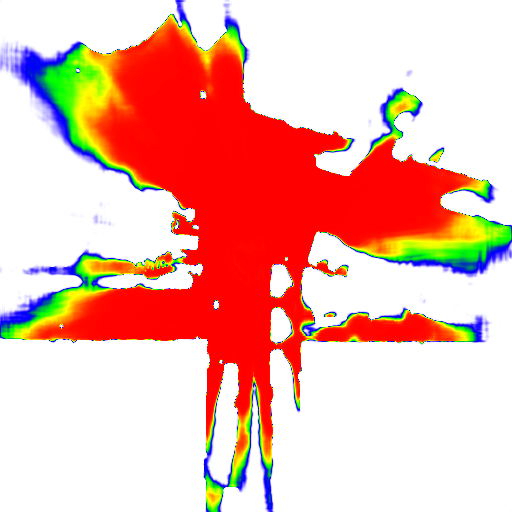}
\caption{Res4}
\label{fig:results_res4}
\end{subfigure}
\begin{subfigure}[h]{0.24\linewidth}
\includegraphics[width=\linewidth]{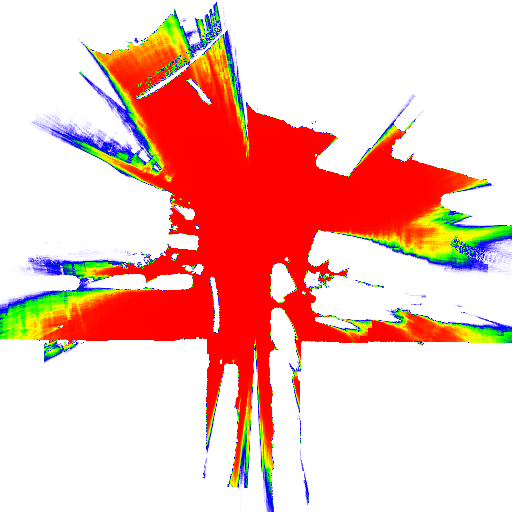}
\caption{Target data}
\label{fig:results_target_data}
\end{subfigure}
\caption{Top row: Grid map layers used as inference network inputs.
Fig.~\ref{fig:results_res4}, \ref{fig:results_u6} and \ref{fig:results_u7} depict the inference output for different networks, where bel($O$) is shown in the middle row and bel($F$) in the bottom row.
Fig.~\ref{fig:results_target_data} depicts the target occupancy grid map used for training the networks.
Low values indicated by white color, high values by red color.
Due to the mapping process, moving objects (e.g. a moving car in the upper left corner) yield high uncertainty in the target data.
However, for a given sensor data frame the network generalizes well and augments the map correctly.
Best viewed digitally with zoom.
}
\label{fig:qualitative_results}
\end{figure*}

We observed that our networks trained with $L_2$ loss yield higher false occupied errors compared to networks trained with $L_1$ loss.
The qualitative inference results depicted in Fig.~\ref{fig:qualitative_results} show that our networks generalize well when moving obstacles are present in the target data.
Compared with input grid map layers split as ground and non-ground, \textit{U5} achieved only a slightly worse performance which shows that the network filters ground surface points to some extent.
Network \textit{U6} achieved comparably high false free/occupied errors but is well suited for grid map augmentation due to its low relative uncertainty.
Network \textit{U7} achieved small false free/occupied errors but therefore has a higher relative uncertainty which might be a gain for accurate grid map filtering instead of augmentation.

\section{Conclusion}
\label{sec:conclusion}

We presented a framework for evidential grid map augmentation using Deep Learning techniques.
By performing quantitative and qualitative evaluation for different configurations, we show that Resnets and Unets infer evidences accurately from domain-specific training data and can be tuned towards grid map augmentation or reliable filtering w.r.t. safety metrics.
Whereas Resnets achieve more accurate results, Unets have a significantly smaller inference time.

In future works we are going to improve the label data generation, especially the registration of range sensor measurements.
This step should decrease target data uncertainty and thus speed up the training process.
Also, we are going to extend our approach to time series of occupancy grid maps in order to augment and predict moving obstacles.

\bibliographystyle{IEEEtran}
\bibliography{root}

\end{document}